\documentclass[10pt,twocolumn,letterpaper]{article}

\usepackage{iccv}              % To produce the CAMERA-READY version
\usepackage{algorithm}
\usepackage{algorithmic}
\usepackage{amsmath}
\usepackage{colortbl}
\definecolor{Row}{RGB}{220,242,230}
\definecolor{Head}{RGB}{210,230,239}
\definecolor{Row2}{RGB}{210,222,230}
\usepackage{makecell}
\usepackage{graphicx}
\usepackage{amssymb}
\usepackage{mathabx}
\usepackage{pifont}
\usepackage{overpic}
\usepackage{tabularx}
\definecolor{iccvblue}{rgb}{0.21,0.49,0.74}
\usepackage[pagebackref,breaklinks,colorlinks,allcolors=iccvblue]{hyperref}

\def\paperID{14253} % *** Enter the Paper ID here
\def\confName{ICCV}
\def\confYear{2025}

\title{Coherent Video Inpainting Using Optical Flow-Guided Efficient Diffusion}

\author{Bohai Gu$^{1,2*}$\;\;\;\;\; Hao Luo$^{2,\dagger}$\;\;\;\;\; Song Guo$^{1,\dagger}$\;\;\;\;\;Peiran Dong$^{1}$\;\;\;\;\; Qihua Zhou$^{1}$\\
$^1$ Hong Kong University of Science and Technology $^2$ DAMO Academy, Alibaba Group \\ \\
\textit{\href{https://nevsnev.github.io/FloED/}{https://FloED.github.io}}
}

\begin{document}

\let\oldtwocolumn\twocolumn
\renewcommand\twocolumn[1][]{
    \oldtwocolumn[{#1}{
    \begin{center}
    \includegraphics[width=\linewidth]{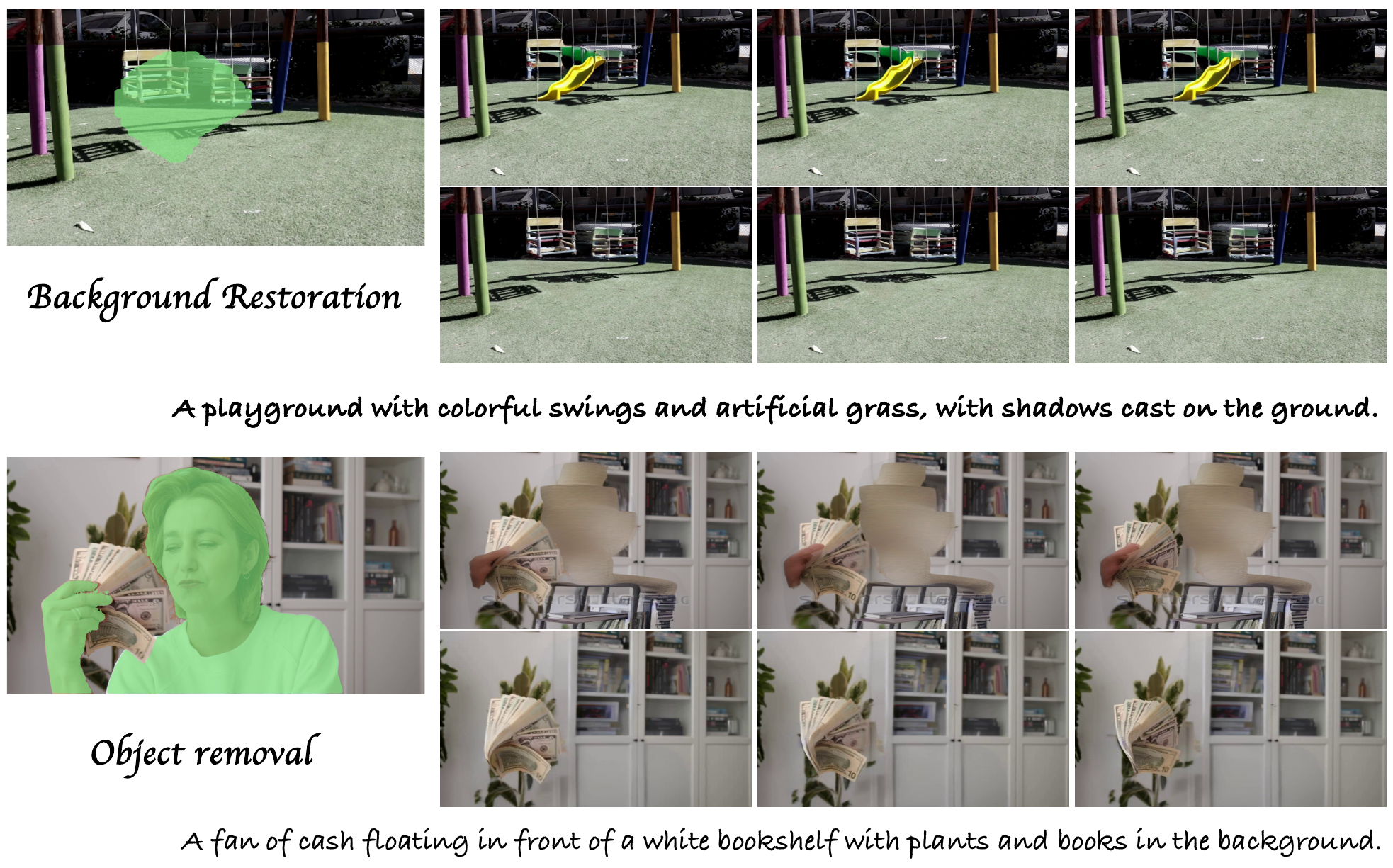}
    \setlength{\abovecaptionskip}{-0.05cm}
    \vspace{-2mm}
    \captionof{figure}{Superiority of FloED in OR and BR tasks is illustrated, with the upper column depicting CoCoCo's results and the lower column highlighting FloED's results.}
    \label{fig:headfig}
    \end{center}
    }]
}

\maketitle

\let\thefootnote\relax\footnotetext{*Work during DAMO Academy internship. $\dagger$ Co-corresponding author.} 
\begin{abstract}

The text-guided video inpainting technique has significantly improved the performance of content generation applications. 
A recent family for these improvements uses diffusion models, which have become essential for achieving high-quality video inpainting results, yet they still face performance bottlenecks in temporal consistency and computational efficiency.
This motivates us to propose a new video inpainting framework using optical \textbf{\underline{Flo}}w-guided \textbf{\underline{E}}fficient \textbf{\underline{D}}iffusion (FloED) for higher video coherence. 
Specifically, FloED employs a dual-branch architecture, where the time-agnostic flow branch restores corrupted flow first, and the multi-scale flow adapters provide motion guidance to the main inpainting branch.
Besides, a training-free latent interpolation method is proposed to accelerate the multi-step denoising process using flow warping. 
With the flow attention cache mechanism, FLoED efficiently reduces the computational cost of incorporating optical flow.
Extensive experiments on background restoration and object removal tasks show that FloED outperforms state-of-the-art diffusion-based methods in both quality and efficiency. Our codes and models will be made publicly available.
\end{abstract}

\section{Introduction}
\label{sec:intro}

Text-guided video inpainting aims to predict corrupted regions with text-aligned and temporally coherent contents, which has drawn widespread attention and applications. 
Conventional transformer-based methods~\cite{DBLP:journals/tip/EbdelliMG15,DBLP:journals/corr/NewsonAFGP15,DBLP:conf/cvpr/0031LQGC22,DBLP:conf/eccv/ZhangFL22,propainter} fundamentally lack textually semantic controllability when users require precise semantic alignment between textual descriptions and generated video results.

In recent years, diffusion models~\cite{DBLP:conf/iclr/SongME21,DBLP:conf/nips/HoJA20,DBLP:conf/cvpr/RombachBLEO22,DBLP:journals/corr/abs-2307-01952} have demonstrated extraordinary capability in generating realistic and text-aligned content. Thus, diffusion-based solutions~\cite{BrushNet,smartbrush,A_Task_is_Worth_One_Word} have made substantial progress in the realm of text-guided image inpainting. However, directly applying these methods to video inpainting falls short in maintaining the necessary temporal consistency.
Although rapid developments in text-to-video (T2V) generative diffusion models~\cite{DBLP:conf/iclr/0002YRL00AL024,tuneavideo} have successfully facilitated text-guided video inpainting~\cite{AVID,CoCoCo,DiffuEraser} by leveraging motion modules~\cite{DBLP:conf/iclr/0002YRL00AL024} for temporal consistency,
it still remains an area with substantial scope for further improvement. 
Meanwhile, current diffusion-based solutions struggle to produce satisfactory results in both Background Restoration (BR) and Object Removal (OR) scenarios. 
Both BR and OR need to generate background content semantically aligned with textual descriptions, while ensuring spatial-temporal coherence between synthesized content and contextual video semantics.

As evidenced in Fig.~\ref{fig:headfig}, state-of-the-art methods like CoCoCo~\cite{CoCoCo} exhibit spatial-temporal disharmony where inpainted regions demonstrate incompatible texture patterns or lighting conditions with surrounding contexts.
We further hypothesize that this disharmony can be mitigated by integrating motion guidance that aligns with the scene. 
In this regard, optical flow, a key modality for capturing motion information, provides valuable guidance and enhances temporal consistency, potentially alleviating the observed inconsistencies.
Notably, integrating optical flow into video inpainting requires additional operations, including flow estimation and flow completion and effective incorporation, all of which incur extra computational costs.
Thus, diffusion models inherently suffer from efficiency due to the multi-step denoising process. Consequently, when we leverage optical flow, it is essential to consider efficiency enhancements tailored to the multi-step characteristics of diffusion models.

Based on the above analysis, we propose a coherent video inpainting framework using optical \textbf{Flo}w-guided \textbf{E}fficient \textbf{D}iffusion, called FloED. 
By leveraging motion information, our approach seeks to enhance both the performance and efficiency of diffusion-based techniques in BR and OR applications. 
Based on the Animatediff~\cite{DBLP:conf/iclr/0002YRL00AL024}, we fine-tune the motion module in the first stage to effectively align its temporal modeling capacity with the video inpainting task.
Built on this primary inpainting branch, our innovations focus on three key aspects:
(1) We design a time-agnostic flow branch that completes corrupted flow while maintaining consistent channel numbers with the primary branch. Next, we integrate the multi-scale flow adapters that inject flow features into the decoder blocks of the primary U-Net architecture, enabling FloED to utilize motion information more effectively. 
% Additionally, we demonstrate that the incorporation of optical flow in this design effectively mitigates artifact hallucinations.
(2) Building on the observation that adjacent latent features share similar motion patterns~\cite{DBLP:conf/eccv/ZhangFL22} and diffusion models fundamentally involve multi-step sampling processes, we introduce a training-free latent interpolation technique, which leverages warping operation guided by optical flow to effectively accelerate the multi-step denoising process during early denoising stage.
Furthermore, by incorporating the flow attention cache mechanism during the remain denoising stage as a complementary speed-up solution, we efficiently minimize the additional computational burden typically introduced by flow adapters and the flow branch.
(3) Recognizing that state-of-the-art image inpainting models significantly outperform video inpainting diffusion models, we utilize an anchor frame strategy to enhance the quality of video inpainting outcomes.

Currently, there is no comprehensive benchmark for evaluating diffusion-based generative approaches in video inpainting. This deficiency presents a substantial challenge, as it limits the ability to rigorously assess and compare the efficacy of various inpainting methodologies. To bridge this gap,  we have developed an extensive benchmark that meticulously encompasses both BR and OR tasks.
Our main contributions are as follows:
\begin{itemize}
    \item \textbf{Novel video inpainting model.} We propose a dedicated dual-branch architecture that integrates optical flow guidance through flow adapters, thereby enhancing spatial-temporal consistency with compatible outcomes.
    \item \textbf{Efficient denoising process.} We introduce a training-free latent interpolation technique that leverages optical flow to speed up the multi-step denoising process. Complemented by the flow attention cache mechanism, FloED efficiently reduces the additional computational costs introduced by the flow.
    \item \textbf{State-of-the-art performance.} We conducted extensive experiments on OR and BR tasks, including both quantitative and qualitative evaluations, to validate that FloED outperforms other state-of-the-art text-guided \textbf{diffusion} methods in terms of both performance and efficiency.
\end{itemize}

\begin{figure*}[t] 
    \centering 
\includegraphics[width=0.99\textwidth]{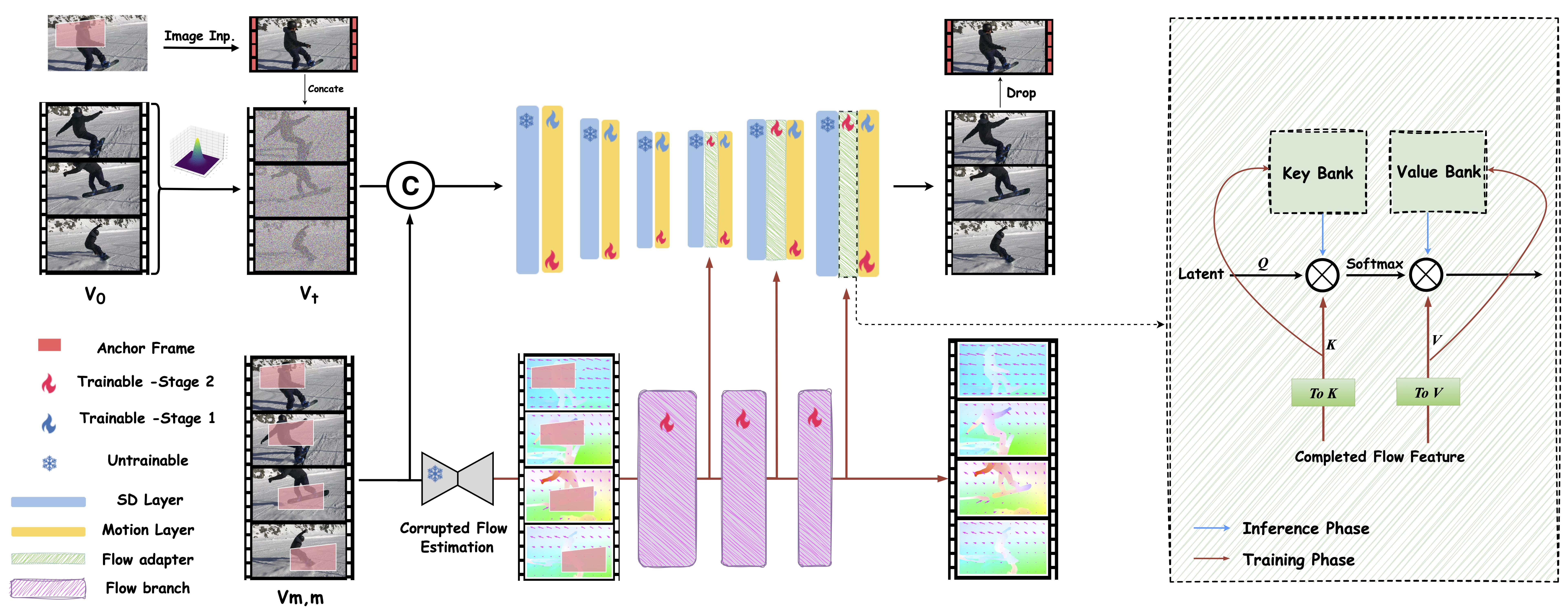}
   % \vspace{-3mm}
    \caption{Overview of FloED. FloED employs a dual-branch architecture implemented through a two-stage training approach. In the first training stage, we focus exclusively on the upper branch, optimizing the motion layer to adapt specifically to the video inpainting domain. Subsequently, we introduce a time-agnostic flow branch complemented by a multi-scale flow adapter, which provides flow guidance covering upblocks of primary UNet. During the inference phase, we enhance efficiency by integrating the flow attention cache (right part). 
    }
    % \vspace{-5mm}
    \label{fig:Overall} 
\end{figure*}

\section{Related Work}
\label{sec:Related Work}

% Definition
Text-guided video inpainting, an emerging topic for synthesizing semantically consistent and temporally stable content in corrupted video regions through natural language instructions, has garnered significant research interest~\cite{AVID,CoCoCo,DiffuEraser} and immense potential for creative and practical industrial applications. 
Although conventional transformer-based solutions~\cite{DBLP:conf/cvpr/0031LQGC22,DBLP:conf/eccv/ZhangFL22,propainter} have shown significant improvements on standard benchmarks DAVIS~\cite{DBLP:conf/cvpr/PerazziPMGGS16} and Youtube-vos~\cite{DBLP:conf/eccv/XuYFYYLPCH18}, they can't provide semantic controllability when users require precise semantic alignment between textual descriptions and generated video results.

In recent years, diffusion models~\cite{DBLP:conf/iclr/SongME21,DBLP:conf/nips/HoJA20} have revolutionized the field of content generation, showcasing exceptional abilities in creating realistic outputs~\cite{DBLP:conf/cvpr/RombachBLEO22,DBLP:journals/corr/abs-2307-01952} and controllable outcomes~\cite{IP-Adapter,Controlnet}. 
The application of diffusion models to text-guided image inpainting has seen remarkable advancements~\cite{BrushNet,smartbrush,A_Task_is_Worth_One_Word}.
However, directly applying these methods to video inpainting falls short in maintaining the necessary temporal consistency.
The emergence of text-to-video technologies~\cite{DBLP:conf/iclr/0002YRL00AL024,tuneavideo,VideoCrafter1}, has provided the solutions for the text-guided video inpainting. 
Notably, Animatediff~\cite{DBLP:conf/iclr/0002YRL00AL024} preserves temporal coherence in video generation through motion module training while effectively utilizing Stable Diffusion's robust framework~\cite{SDv15}. 
Building upon this foundation, recent approaches~\cite{AVID,CoCoCo,DiffuEraser} have advanced text-guided video inpainting. Specifically, AVID~\cite{AVID} proposes a fine-grained generation pipeline with structure guidance. CoCoCo~\cite{CoCoCo}  employs enhanced attention mechanisms to improve text-video alignment and motion consistency. And DiffuEraser~\cite{DiffuEraser} introduces the prior model for better inpainting outcomes.
Nevertheless, existing methods demonstrate suboptimal performance on both object removal (OR) and background restoration (BR) tasks, frequently producing incompatible texture patterns or illumination conditions with surrounding contexts. In comparison, FloED exhibits superior temporal consistency and overall output quality.

\section{Preliminaries}
LDM~\cite{DBLP:conf/cvpr/RombachBLEO22} leverages a pre-trained Variational Autoencoder (VAE) to operate in the latent space instead of pixel space. 
The diffusion forward process is imposing noise on a clean latent $\mathbf{z}_0$ for $T$ times.
A property of the forward process is that it admits sampling $\mathbf{z}^t$ at random timestep $t$:
\begin{equation}\label{equ: to noise}
     q(\mathbf{z}^t | \mathbf{z}^0) = \mathcal{Q}(\mathbf{z}^0, t) = \mathcal{N}(\mathbf{z}^t; \sqrt{\bar{\alpha}_t} \mathbf{z}^0, (1-\bar{\alpha}_t) \mathbf{I}),
\end{equation}
where $\bar{\alpha}_t=\prod_{s=1}^t (1 - \beta_s)$, $\beta_s$ is the variance schedule for the timestep $s$.
% and we use $\mathcal{Q}(\cdot,\cdot)$ to represent this one-step noising process. 
The backward process applies a trained UNet $\epsilon_\theta$ for denoising: $p_\theta(\boldsymbol{z}_{t-1}|\boldsymbol{z}_t)=\mathcal{N}(\boldsymbol{z}_{t-1};\mu_\theta(\boldsymbol{z}_t,t),\Sigma_\theta(\boldsymbol{z}_t,t)),$
where distribution parameters $\mu_\theta$ and $\Sigma_\theta$ are computed by the denoising model $\theta$.
To train a conditional LDM, the objective is given by:
\begin{align}
\label{euq:objective}
\mathcal{L}_{\textrm{diff}} = \underset{\theta}{\mathrm{arg\,min}} \; 
\mathbb{E}_{\mathbf{z},\epsilon\thicksim\mathcal{N}(0,1),t,c}\left[\|\epsilon-\epsilon_\theta(\mathbf{z}_t,t,c)\|_2^2\right],
\end{align}
where $\epsilon_\theta(\mathbf{z}_t,t,c)$ is the predicted noise based on $\mathbf{z}_t$, the time step $t$ and the condition $c$. Once trained, we could leverage the deterministic sampling of DDIM~\cite{DBLP:conf/nips/HoJA20} to denoise $\mathbf{z}_t$:

\begin{equation}
\label{euq: ddim}
\begin{aligned}
\mathbf{z}_{t-1} &= \sqrt{\alpha_{t-1}} \underbrace{\hat{\mathbf{z}}_{t\rightarrow0}}_{\text{predicted `$\mathbf{z}_0$'}} + \\
&\underbrace{\sqrt{1-\alpha_{t-1} - \sigma_t^2} \epsilon_{\theta}(\mathbf{z}_t, t, c)}_{\text{direction pointing to $\mathbf{z}_t$}} + \underbrace{\sigma_t \epsilon_t}_{\text{random noise}}, 
\end{aligned}
\end{equation}
where $\sigma_t$ are hyper-parameters. 
The term ${\mathbf{z}}^t_{t \rightarrow 0}$ represents the predicted $\mathbf{z}_0$ at time step $t$, 
% which is characterized through the operation \(\mathcal{P}(\cdot,\cdot)\), as delineated in the equation below.
For conciseness and to circumvent any potential confusion with the concept of optical flow, we subsequently refer to \(\hat{\mathbf{z}}_{t\rightarrow0}\) as \(\hat{\mathbf{z}}_0\). The precise formulation is as follows:
\begin{equation}
\label{eq:to_x0}
  \hat{\mathbf{z}}_{t\rightarrow0}\ = \mathcal{P}(\mathbf{z}_t, \epsilon_\theta) = (\mathbf{z}_t-\sqrt{1-\alpha_{t}}\epsilon_\theta(\mathbf{z}_t, t, c))/\sqrt{\alpha_{t}}.
\end{equation}

\section{Methods}
\label{sec:Methods}

\begin{figure*}[t] 
    \centering 
\includegraphics[width=0.95\textwidth]{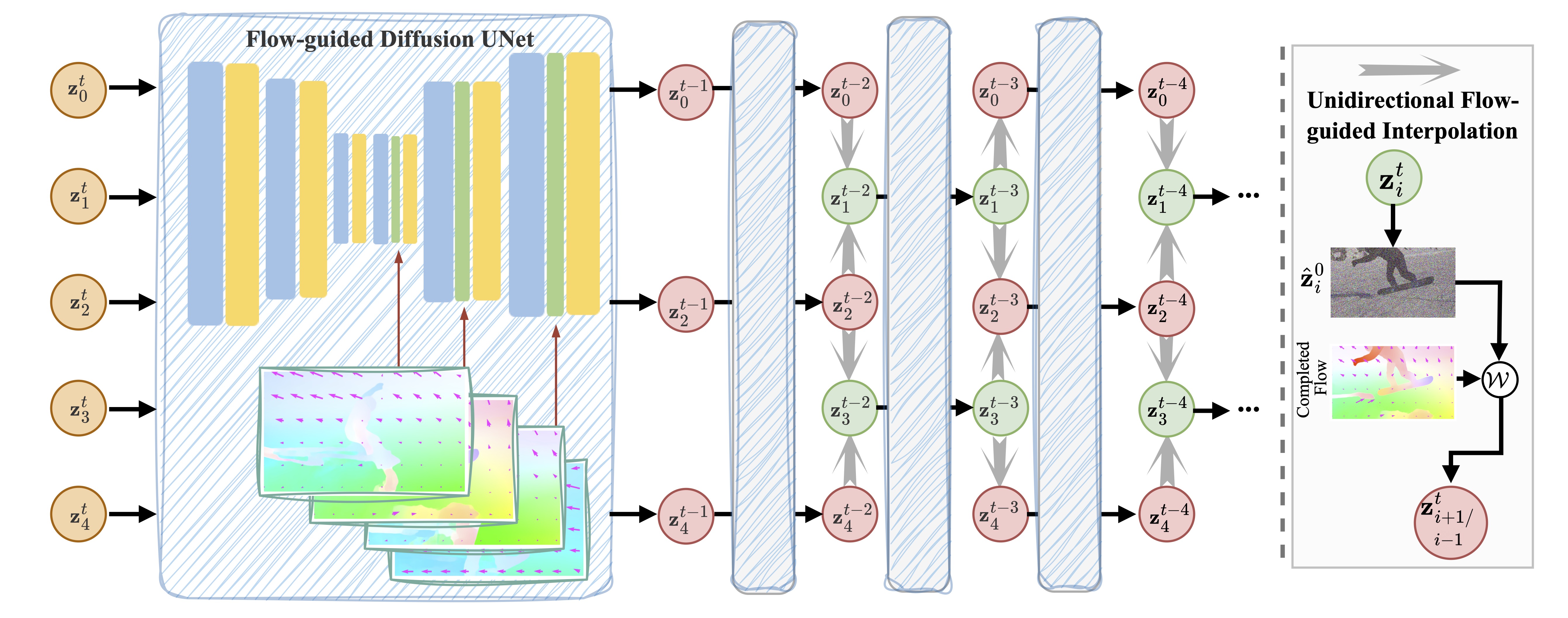}
   % \vspace{-3mm}
    \caption{Illustration of flow-guided latent interpolation (left) and warping operation (right) during the denoising process.}
       % \vspace{-5mm}
    \label{img:Acceleration} 
\end{figure*}
Given a text prompt $P$ and original video sequence \(\mathbf{V_0} = \{\mathbf{x}_0, \mathbf{x}_1, \ldots, \mathbf{x}_{N-1}\}  \in \mathbb{R}^{N \times 3 \times H \times W}\) and a binary mask sequence \(\mathbf{m} = \{\mathbf{m}_0, \mathbf{m}_1, \ldots, \mathbf{m}_{N-1}\} \in \mathbb{R}^{N \times 1 \times H \times W}\), corrupted frames  $\mathbf{V_m}$ are obtained by applying the Hadamard product as follows: \(\mathbf{V_m} = \mathbf{V_0} \odot \mathbf{m}\). We aim to generate a set of spatiotemporally consistent inpainted outcomes with text-aligned contents which demonstrate compatible texture patterns and lighting conditions with surrounding context.
\subsection{Network Overview}
An overview of our proposed model, FLoED, is depicted in Fig.~\ref{fig:Overall}. For the architecture, FloED adopt pretrained Stable Diffusion Inpainting backbone~\cite{SDv15} as the primary branch, while incorporating motion modules initialized from AnimateDiff v3~\cite{DBLP:conf/iclr/0002YRL00AL024}. The training process of FloED can be divided into two stages. 

In the first stage, we fine-tune the motion modules to align their temporal modeling capacity with video inpainting.  
In the second stage, FloED incorporates a dedicated flow branch to complete corrupted flows estimated from masked frames, alongside multi-scale flow adapters that inject hierarchical motion information into the primary inpainting branch (Sec.~\ref{sec:Flow Guidance in video inpainting}). 
To further enhance video inpainting results, we implement an anchor frame strategy to leverage the priority of the image inpainting diffusion model~\cite{SDv2} (Sec.~\ref{sec:Anchor Frame Strategy}). 
Furthermore, we introduce a training-free denoising acceleration technique that leverages optical flow for latent interpolation,  compensated with a flow attention caching mechanism in the reference phase. We substantially enhance efficiency while significantly reducing the additional computational overhead the flow introduces. (Sec.~\ref{sec:Efficient Inference for FloED}).

Notably, noised video latent $\mathbf{V_t}$, down-sampled binary mask $\mathbf{m}$, 
corrupted video latent $\mathbf{V_m}$ are concatenated along the channel axis as input.

\subsection{Flow Guidance in Video inpainting}
\label{sec:Flow Guidance in video inpainting}

Considering the flow guidance, FloED first employs a pre-trained flow generator RAFT~\cite{DBLP:conf/ijcai/Teed021} to obtain the corrupted flow from $\mathbf{V_m}$, which is annotated as "Corrupted Flow Estimation" in Fig.~\ref{fig:Overall}.

\paragraph{Flow Completion Branch.}
To reconstruct the corrupted flow, the framework incorporates a dedicated flow completion branch that architecturally aligns with the primary inpainting backbone. Specifically, it selectively aggregate the initial ResNet module~\cite{resnet} from each corresponding block in the primary branch, to ensure channel-wise compatibility. And we exclude the time-step from ResNet to obtain the time-agnostic flow completion branch which creates stable flow feature regardless of diffusion progression. As shown in Fig~\ref{fig:Overall}, this flow completion branch enables comprehensive motion guidance by injecting compatible motion information into up-blocks of the primary UNet branch with multi-scale flow adapters.

\paragraph{Flow Adapter.}
Our flow adapter is inspired by IP-Adapter~\cite{IP-Adapter}, which is consisted with a separated cross-attention layer. The reconstructed flow features are fed into the cross attention for motion guidance.
Notably, FloED strategically positions the flow adapter between the text cross-attention and motion modules to enable flow-aware latent modulation. This critical design further addresses misalignment caused by text-driven generation, because the multi-scale flow adapters dynamically adjusts latent features from the text cross-attention layer using optical flow priors, ensuring synthesized content maintains spatial-temporal compatibility with surrounding contexts.
And in the second stage, we also continually fine-tune the motion modules.

The reconstructed optical loss $\mathcal{L}_{\textrm{flow}}$ is introduced in the second stage. ($F$ represents the ground truth flow)
\begin{equation}\label{euq:loss}
    \mathcal{L}_{\textrm{sec}} = \mathcal{L}_{\textrm{diff}} +   \lambda*\mathcal{L}_{\textrm{flow}},\mathcal{L}_{flow}= \|\hat{F}-F\|_{1} 
\end{equation}

\subsection{Anchor Frame Strategy}
\label{sec:Anchor Frame Strategy}
Recognizing that state-of-the-art image inpainting models developed in academia far surpass their video inpainting counterparts, we introduce an anchor frame strategy that leverages these advanced image techniques to substantially improve video inpainting performance.
As illustrated in Fig.~\ref{fig:Overall}, for a given video sequence $\mathbf{V_0}$, we select an additional frame from the beginning of the sequence to serve as an anchor frame. We then utilize a pre-trained text-to-image (T2I) inpainting model~\cite{SDv2} to reconstruct its corrupted region in advance.
Subsequently, we concatenate the inpainted anchor frame with the noised video frames $\mathbf{V_t}$. This approach provides additional texture guidance to the video frames during the denoising process.
After denoising, the anchor frame is discarded. This strategy leverages the superior performance of image inpainting models to improve the overall quality of video inpainting.

\subsection{Efficient Inference for FloED}
\label{sec:Efficient Inference for FloED}

Based on multi-step sampling processes of diffusion, we further propose a training-free latent interpolation technique that leverages optical flow to speed up the denoising process. This approach is complemented by a flow attention cache mechanism during the inference phase.

\paragraph{Flow Attention Cache.} 
% first step flow only 
Unlike the primary branch, the flow branch is independent of timestep. During the inference phase, we utilize the flow branch exclusively for flow completion in the first step and then use these completed flows for all subsequent steps. Regarding the multi-scale flow adapters, they introduce additional computations by calculating the flow attention at every denoising step and multiple resolutions. To optimize this process, we establish a cache mechanism by computing the keys and values only during the first step and storing them in the memory bank (right part of Fig.~\ref{fig:Overall}). For the remaining steps, the cached keys and values are directly retrieved from the memory bank, eliminating the need for repeated calculations and enhancing efficiency.

\paragraph{Training-Free Denoising Speed-up.} 
Since adjacent feature latent exhibit similar motion patterns~\cite{DBLP:conf/eccv/ZhangFL22} and diffusion models generate high-level content early in the denoising process~\cite{DBLP:conf/iclr/KwonJU23}, we aim to speed-up the denoising process by interpolating latent using the completed flow. Notably, this technique is entirely training-free.

Specifically, as illustrated in Fig.~\ref{img:Acceleration}, the initial step involves performing the standard denoising process for completing flow and caching flow attention. Subsequently, starting from step $t-1$, the noisy latent $\mathbf{z}$ is divided into two subsets based on parity. The latent interpolation process then follows a two-step alternating loop: even-indexed latents (shown in red) undergo denoising, while odd-indexed latents (shown in green) are obtained by warping operations using bi-directional optical flows. In the next step, only the interpolated latents (green) are denoised, and the red latents are generated through a similar warping process. Due to the negligible time cost of warping latent, the latency of denoising is halved by processing only half of the frame latent at each sampling timestep. Notably, the warping operation needs to be conducted at 
$\mathbf{z}_0$ (as per Equation~\ref{eq:to_x0}). 

\begin{figure*}[t] 
\includegraphics[width=0.99\textwidth]{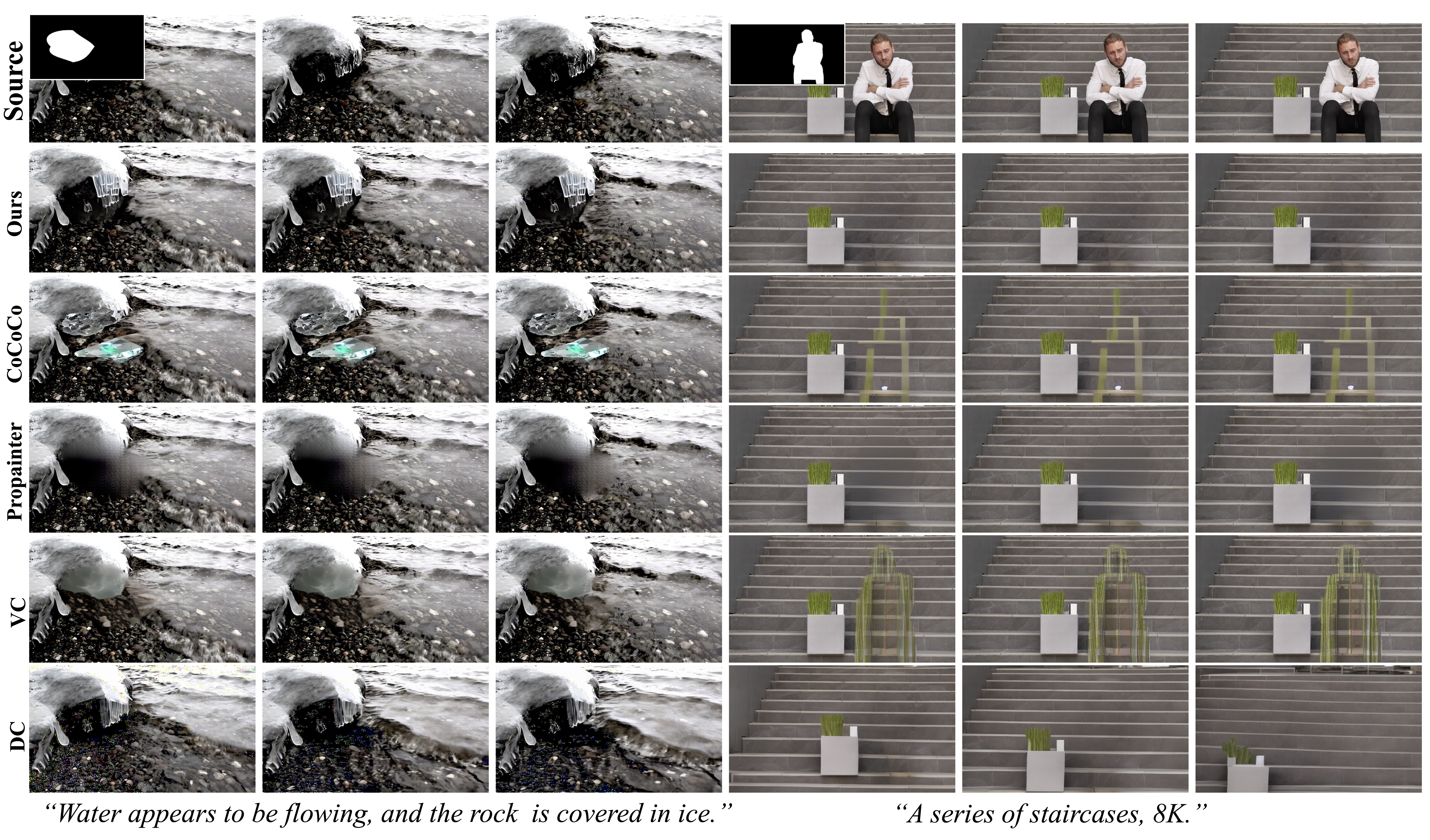}
\vspace{-2mm}
\caption{\textbf{Qualitative comparisons}. We compare FloED against diffusion-based SOTAs on BR and OR tasks. }
% \vspace{-5mm}
\label{fig:Qualitative comparisons.}
\end{figure*}

Ideally, this process could be iterated until the final denoising step. However, since we are operating in latent space, completed optical flow provides only coarse-grained guidance, corresponding to the early stage of the diffusion process. Therefore, we restrict latent interpolation to the initial $S$ denoising steps, during which the overall structure of the image is established~\cite{DBLP:conf/iclr/KwonJU23}.
Additionally, to minimize flow errors, we perform warp operations exclusively between adjacent frames.
Furthermore, to mitigate potential occlusion issues in flow warping, we perform the copy-paste operation in each denoising step (see in appendix).

\section{Experiments}
\label{sec:Methods}

\begin{table}[t]
    \centering
    \resizebox{0.5\textwidth}{!}{ % 调整表格宽度为一半
        \begin{tabular}{l l l l c l l}
        \toprule
        & \multicolumn{5}{c}{BR} & \multicolumn{1}{l}{OR}  \\
        \cmidrule(lr){2-6} \cmidrule(lr){7-7}
        & PSNR ↑ & SSIM ↑  & VFID ↓ &${E_{warp}}$↓ & TC ↑ & TA ↑  \\
        \midrule
        VC       & 22.81  & 0.8614 & 0.193  & 3.43  & 0.987  & 21.30 \\
        CoCoCo   & 23.08  & 0.8694 & 0.165  & 3.73  & 0.991  & 21.97 \\
        DiffuEraser & 24.23 & 0.8583 & 0.218 & 2.98  & 0.984  & 19.19     \\
        Ours     & \bf{29.17} & \bf{0.9441} & \bf{0.118} & \bf{2.83} & \bf{0.994} & \bf{22.49} \\
        \bottomrule
        \end{tabular}
    }
    \caption{
        \textbf{Quantitative comparisons.} 
        Metrics including PSNR, SSIM, VFID, Optical flow warping error(${E_{warp}}$), 
        Temporal consistency (TC), and Text alignment (TA).
    }
    \label{table: Quantitative comparison}
\end{table}

\subsection{Implementation Details.}
\paragraph{Dataset and Benchmark.}
We utilize the Open-Sora-Plan dataset~\cite{pku_yuan_lab_and_tuzhan_ai_etc_2024_10948109}, splitting videos at scene cuts to obtain 421,396 high-quality video clips paired with captions. We further developed an evaluation benchmark comprising 100 previously unseen videos sourced from Pexels and Pixabay platforms, with 50 designated for object removal (OR) and 50 for background restoration (BR). For BR task, we use synthetic random masks that focus on the background. For the OR task, object masks are obtained by applying Segment-Anything~\cite{kirillov2023segment} (SAM) to each frame. Each video is manually selected to ensure a diverse range of motion amplitudes and camera movement speeds, while guaranteeing 4K resolution and 100 frames in total. For the caption, we generate the corresponding video prompts using VideoGPT~\cite{Maaz2023VideoChatGPT}, while manually revising appropriate background prompts for OR.

\paragraph{Training and Inference Details.}

We employed a two stage training strategy with a resolution of 512 and 16-frame video sequences. And we generated mask sequences with random directions, and shapes to simulate BR and OR tasks. The first stage is trained on 8 NVIDIA A800 GPUs for 5 epochs with a batch size of 8. And second stage is trained on 8 NVIDIA A800 GPUs for 30 epochs with a batch size of 128, which is achieved through gradient accumulation the $\lambda$ is set to 0.1 during the second stage. During inference, we use DDIM~\cite{DBLP:conf/nips/HoJA20} and empirically define the speed-up step $S$ as 5 (25 steps in total).

\subsection{Comparisons}
We present comprehensive comparisons with open-sourced \textbf{text-guided} diffusion-based approaches, including VideoComposer~\cite{VideoComposer}, CoCoCo~\cite{CoCoCo} and DiffuEraser~\cite{DiffuEraser}.

\paragraph{Qualitative Comparisons.}
As demonstrated in Fig.~\ref{fig:Qualitative comparisons.}, VideoComposer, CoCoCo and DiffuEraser exhibit persistent limitations especially in OR tasks, frequently generating visual artifacts and content hallucinations that disrupt semantic consistency with scene context. In contrast, FloED inpaints the mask region with compatible contents and demonstrates precise text-conditioned generation capabilities, achieving superior temporal consistency and overall coherence in both BR and OR tasks.

\begin{figure*}[t] 
\centering 
\includegraphics[width=\textwidth]{Figure/Flow_ablation.jpg}
% \vspace{-2mm}
\caption{Optical flow related ablation studies. (F) ablation study demonstrates FloED conducts flow warping at noise $\epsilon$ instead  of $\mathbf{z}_0$.} 
% \vspace{-3mm}
\label{fig:flow related ablation.}
\end{figure*}

\begin{table}[t]
\centering
\small
    \begin{center}
    \centering
    \begin{tabular}{ll|ccccc}
    % \cline{1-7} 
\toprule
\bf{FA}  & \bf{AF}  & PSNR ↑ & SSIM ↑  & VFID ↓ &${E_{warp}}$↓ & TC ↑ \\
    % \cline{1-7} 
    % \hline
    % \centering
% \CheckmarkBold  & \CheckmarkBold  &21.05 &0.8402 &0.421 &20.5  &0.994 \\
    \hline
  \ding{55} &  \ding{55}  &21.30 &0.8435 &0.246 &4.15  & 0.989\\
    % \hline
  \ding{55}  & \ding{51}  &25.34 &0.9138 &0.195 &3.87  &0.984 \\
    % \hline
 \ding{51} & \ding{55} &27.05 &0.9255 &0.170 &3.03  &0.989 \\
    % \hline
 \ding{51} & \ding{51} &28.71 &0.9401 &0.125  &2.95 &0.990 \\
    % \hline 
 \ding{51} & \ding{51}  &\bf{29.17}  &\bf{0.9441}  &\bf{0.118} &\bf{2.83}  & \bf{0.994} \\
    \bottomrule
    \end{tabular}
    \end{center}
    \vspace{-5mm}
    \caption{\textbf{Ablation study}. \textbf{FA}, \textbf{AF} stand for Flow Adapter and Anchor frame, respectively. All variants were trained using identical settings, except the last one with extended training (30 epochs).} 
    \vspace{-3mm}
    \label{table:Ablation Study}
\end{table}
\begin{figure}[b] 
    \centering 
\includegraphics[width=0.48\textwidth]{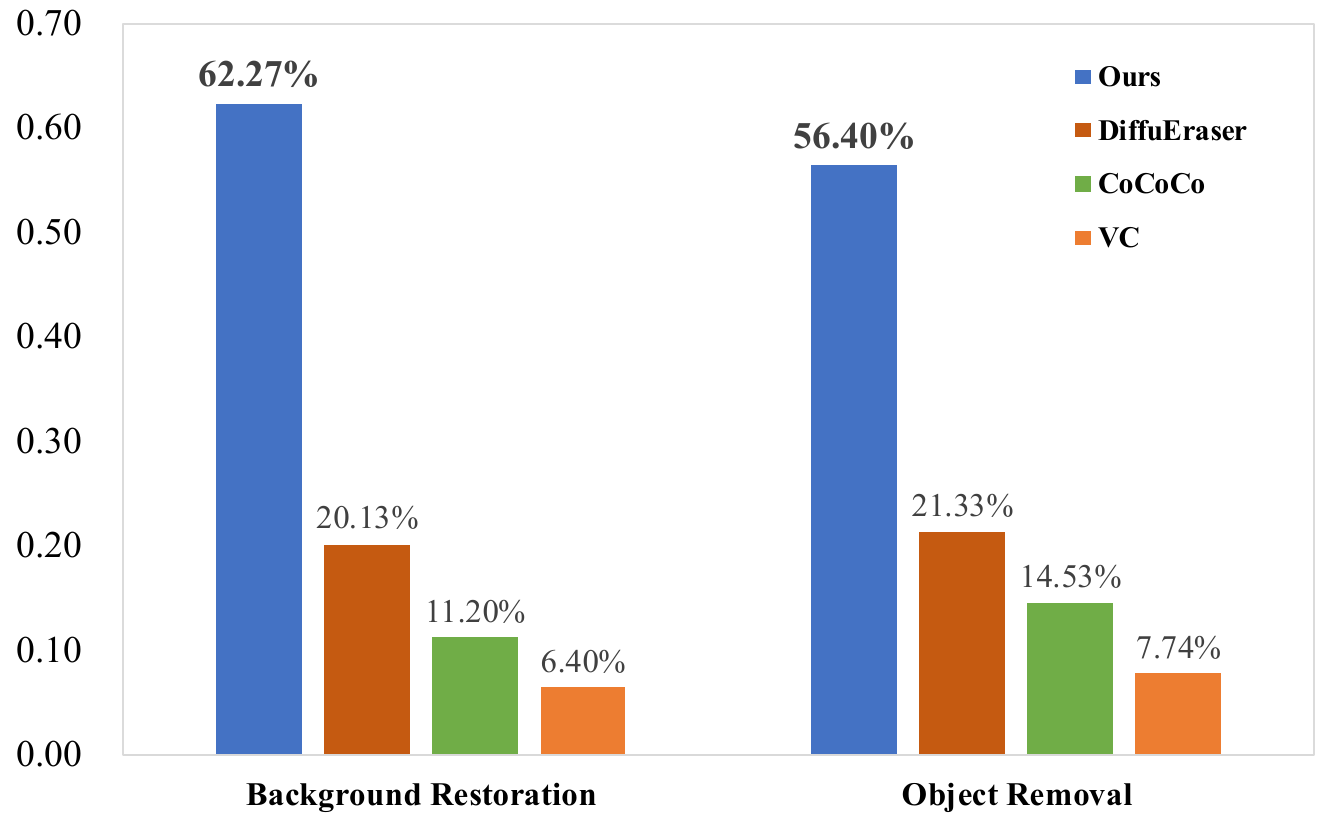}
% \vspace{-5mm}
    \caption{ We conduct reliable \textbf{User Study} with randomized order to assess inpainting outcomes of 4 methods. }
    \label{img:User_study} 
    % \vspace{-3mm}
\end{figure}

\paragraph{Quantitative Comparisons.} We take metric evaluation and user studies to demonstrate quantitative comparisons.

\noindent\textbf{(1) Metric evaluation.} For BR tasks, we employ the PSNR~\cite{PSNR}, VFID~\cite{DBLP:conf/nips/Wang0ZYTKC18}, and SSIM~\cite{DBLP:journals/tip/WangBSS04} to quantify basic quality. Additionally, we assess temporal consistency using flow warping error~\cite{DBLP:conf/iccv/GuFZ23,DBLP:conf/iclr/KwonJU23} in conjunction with Temporal Consistency (TC)~\cite{AVID}. TC is measured by the cosine similarity between consecutive frames in the CLIP-Image~\cite{Clip} feature space.
For OM tasks, since ground truth data is unavailable for evaluating the aforementioned metrics, we utilize Text Alignment (TA)~\cite{Clip} as an evaluation metric, which also leverages the CLIP score. 
For consistency, all metric evaluations were conducted at a resolution of 512×512. As shown in Tab.~\ref{table: Quantitative comparison}, FloED outperforms other methods in all the metrics, which demonstrates state-of-the-art performance.

\noindent\textbf{(2) User study.}
Since CLIP scores do not always align with human perception~\cite{AVID}, we conducted a comprehensive user study in which 15 annotators evaluated the inpainting results across both BR and OR tasks (100 videos), assessing temporal consistency, text alignment, and context compatibility to select the best one among 4 methods.As illustrated in Fig.~\ref{img:User_study}, our model was highly favored, achieving the highest scores in both BR (62.27\%) and OR (56.40\%).

\subsection{Ablation Studies}

\paragraph{Flow-related Ablation Studies.}
We conduct flow-related ablation studies to validate the motion guidance, with experimental results shown in Fig.~\ref{fig:flow related ablation.}. 

\noindent\textbf{(1) Flow completion.} Using an object removal scenario as a case study, completed optical flow results demonstrate that the corrupted flow undergoes context-aware inpainting with spatially and temporally coherent content which maintains alignment with the surrounding environment (comparing B with C). And the reconstructed results further validate the completion capability of our time-agnostic flow branch.

\noindent\textbf{(2) Flow adapter.} These reconstructed flows provide crucial motion guidance to the primary inpainting branch, effectively enhancing context compatibility and achieving higher video coherence through multi-scale flow adapters. Comparing the outcomes of D with E, black artifacts and incompatible lighting condition can be largely reduced. The experimental findings demonstrate that the multi-scale flow adapters' injection of motion guidance significantly improves environmental consistency in generated content, which in turn enhances temporal coherence and overall quality. Meanwhile,  we also conduct the quantitative architecture ablation study  in Tab.~\ref{table:Ablation Study}. Compared with anchor frame strategy, the multi-scale flow adapters demonstrate superior efficacy in enhancing framework's performance, confirming their critical role in FloED.  More ablation study results are provided in the Appendix.

\noindent\textbf{(3) Flow warping operation.} Instructively, our ablation study F reveals that applying flow warping to the intermediate noise estimate $\epsilon$ during latent interpolation  leads to cumulative error propagation with blurriness  results compared to baseline D. To circumvent this,  we reformulate the warping operation in the clean latent space $\mathbf{z}_0$(as per Equation~\ref{eq:to_x0}).

\begin{figure}[t] 
    \centering 
\includegraphics[width=0.50\textwidth]{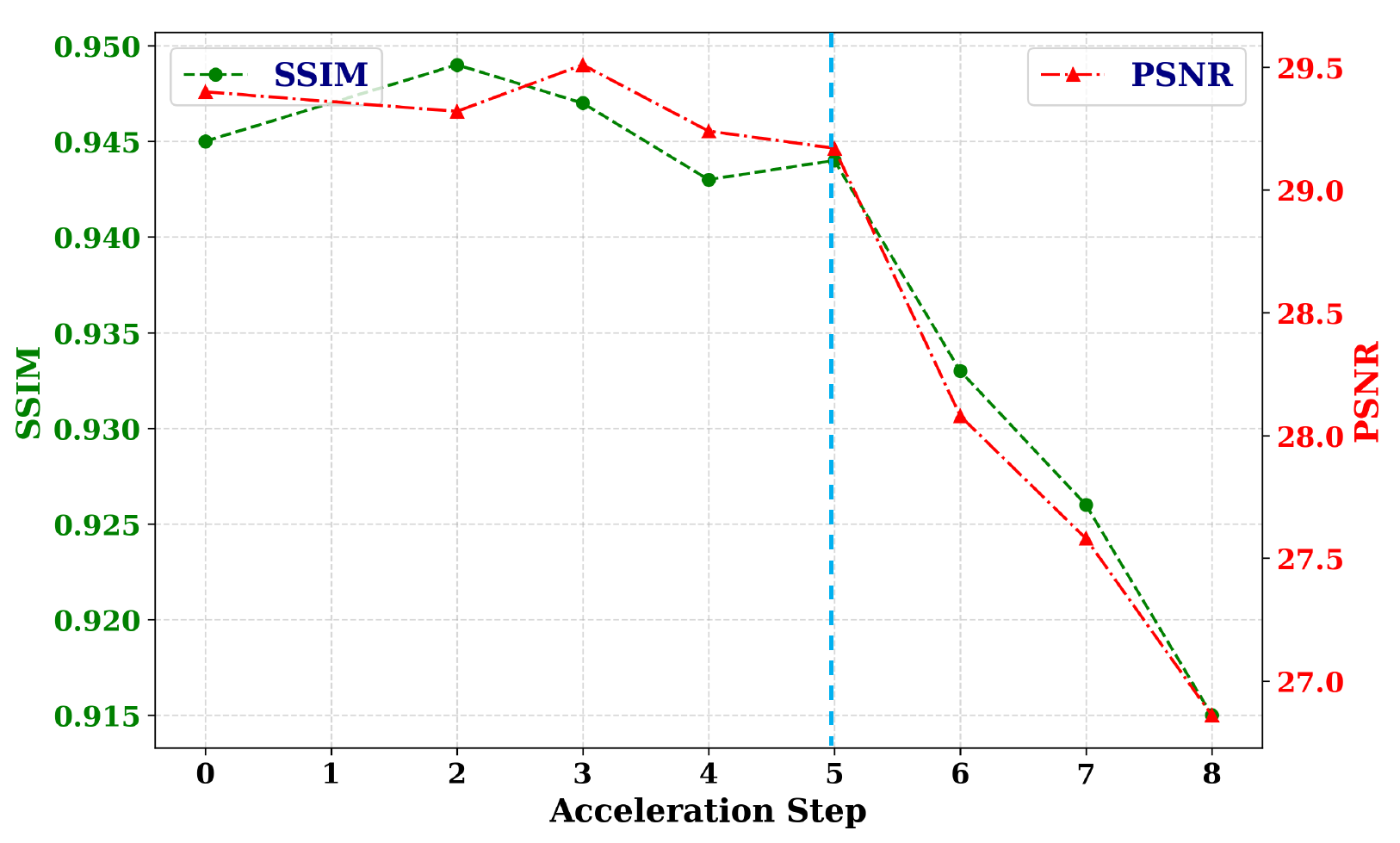}
% \vspace{-5mm}
    \caption{\textbf{Speeding steps study}. Performance markedly deteriorates when the acceleration step surpasses five. }
    \label{Ablation Step} 
    % \vspace{-5mm}
\end{figure}

\begin{table}[t]
    \footnotesize % 缩小字体
    \centering
    \setlength{\tabcolsep}{6pt} % 减少列间距
    \renewcommand{\arraystretch}{1.2} % 增加行高

    \begin{tabularx}{0.47\textwidth}{llll} % 设置表格宽度为单栏的 80%
    \toprule
    % \rowcolor{gray!20} % 表头背景色
    \textbf{Flow Branch} & \textbf{\makecell{Latent \\ Interpolation}} & \textbf{Flow Cache} & \textbf{\makecell{Average time  \\ per frame(s)}}\ \\
    \midrule
    -- & -- & -- & \textcolor{blue}{0.1287} \\
    $1^{st}$ step & -- & -- & 0.1491 \\
    $1^{st}$ step & $2^{nd} \sim 6^{th}$ & -- & 0.1342 \textcolor{red}{(↓ 9.9\%)} \\
    % \rowcolor{gray!10} % 行背景色
    $1^{st}$ step & $2^{nd} \sim 6^{th}$ & $6^{nd} \sim 25^{th}$ & 0.1291 \textcolor{red}{(↓ 13.4\%)} \\
    \bottomrule
    \end{tabularx}

    \caption{\textbf{Efficiency ablation study} (432 $\times$ 240).  The second row represents variant with no flow-related module involved (blue). }
    \label{table:Efficiency analysis}
\end{table}

\paragraph{Efficiency.}
In this section, we conduct the efficiency experiments with single H800 NVIDIA GPU under FP16 setting. And FloED's denoising process is executed for 25 steps with the classifier-free guidance scale (CFG)  $>$  1. 

\noindent\textbf{(1) Latent interpolation steps.}
As discussed in Sec.~\ref{sec:Efficient Inference for FloED}, we apply latent interpolation exclusively during the initial phase. Thus, we conduct a speeding steps study in Fig.~\ref{Ablation Step}. As depicted in Fig.~\ref{Ablation Step}, continuously increasing the acceleration step results in a sharp decline in performance when the speeding steps $S$ beyond the early stage of the denoising process.  Our experimental findings demonstrate that utilizing flow-guided latent interpolation for the initial 5 steps, we can minimize denoising time with only a slight compromise in performance.
 
\noindent\textbf{(2) Efficiency ablation study.}
As presented in Tab.~\ref{table:Efficiency analysis}, we compare different efficiency strategies against the variant without any flow-related module. As previously mentioned, since our flow branch is independent of time-step during training, during the testing phase, we only need to utilize the flow branch in the first step of the denoising process to complete the damaged optical flow and cache the memory bank. 
For the remaining denoising steps, we can directly use the completed flow for latent interpolation and the cached K, V for flow guidance.

Thus, we determined an optimal solution: applying latent interpolation for the initial 5 steps ($2^{nd} \sim 6^{th}$) and leveraging flow caching for the remaining steps for complement, resulting in 13.4\% speed-up in the resolution of 432$\times$240.
Compared to the pure variant, which does not incorporate flow completion and flow attention, these efficiency benefits nearly offset the additional computational burden, incurring minimal cost.

\noindent\textbf{(3) Efficiency comparisons.}
As shown the Tab.~\ref{Tab:C}, under the same denoising steps, FloED outperform other diffusion-based counterparts, such as CoCoCo and DiffuEraser, across all resolutions, demonstrating its state-of-the-art efficiency.

\begin{table}[t]
    \footnotesize % 缩小字体
    \centering
    \setlength{\tabcolsep}{6pt} % 减少列间距
    \renewcommand{\arraystretch}{1.0} % 减少行高

    \begin{tabularx}{0.45\textwidth}{c|ccc} % 设置表格宽度为单栏的 80%
    \toprule
    \textbf{Average time per frame (s)} & \textbf{512$\times$512} & \textbf{432$\times$256} & \textbf{256$\times$256} \\
    \midrule
    \textbf{Ours}        & \textbf{0.3412} & \textbf{0.1291} & \textbf{0.0756} \\
    CoCoCo               & 0.4033          & 0.2084          & 0.1735          \\
    DiffuEraser          & 0.4176          & 0.5751          & 1.0336          \\
    \bottomrule
    \end{tabularx}
  % \vspace{-2mm}
    \caption{\textbf{Efficiency comparisons} (CFG $>$  1, 25 steps, FP16).}
    \label{Tab:C}
  % \vspace{-2mm}
\end{table}

\paragraph{Discussion.}

This paper focuses on text-guided video inpainting, primarily comparing with diffusion-based solutions. Our FloED also demonstrates superior performance over conventional flow-guided methods like ProPainter~\cite{propainter}, with detailed comparisons provided in the Appendix. Furthermore, FloED's latent interpolation can be directly extended to other diffusion-based approaches (e.g., CoCoCo~\cite{CoCoCo}) for accelerated processing. However, we note that the strategy of pre-completing corrupted optical flow might restrict its transferability across different application scenarios.

\section{Conclusion}
\label{sec:Conclusion}
In this paper, we introduced FloED, a coherent video inpainting framework that effectively integrates optical flow guidance into diffusion models to enhance temporal consistency and computational efficiency. By employing a dual-branch architecture, FloED first reconstructs corrupted flow, which then guides the inpainting process through multi-scale flow adapters. Additionally, our training-free latent interpolation technique and flow attention cache significantly reduce the computational overhead typically associated with optical flow integration. Experimental results demonstrate that FloED achieves state-of-the-art performance in both background restoration and object removal, showcasing its superior ability to maintain temporal consistency and content coherence in video inpainting.

{
    \small
    \bibliographystyle{ieeenat_fullname}
    \bibliography{main}
}

\end{document}